\documentclass[final]{agujournal2019}
\usepackage[finalnew]{trackchanges} 

\usepackage{soul}
\usepackage{flafter}
\usepackage{graphicx}
\usepackage{tikz}
\usetikzlibrary{positioning}
\usepackage{subcaption}
\usepackage{forest}
\usepackage{booktabs}
\usepackage{fancyhdr}

\usepackage{url}

\usepackage{amsmath,amsfonts,bm}

\def\1{\bm{1}}

\DeclareMathAlphabet{\mathsfit}{\encodingdefault}{\sfdefault}{m}{sl}
\SetMathAlphabet{\mathsfit}{bold}{\encodingdefault}{\sfdefault}{bx}{n}

\newcommand{\bfA}{{\bf A}}

\newcommand{\bfI}{{\bf I}}

\newcommand{\bfg}{{\bf g}}

\newcommand{\bfx}{{\bf x}}

\newcommand{\bfu}{{\bf u}}

\newcommand{\bfd}{{\bf d}}
\newcommand{\bfm}{{\bf m}}

\newcommand{\bfv}{{\bf v}}
\newcommand{\bfw}{{\bf w}}
\newcommand{\bfz}{{\bf z}}

\newcommand{\grad}{{\boldsymbol \nabla}}

\newcommand{\bfepsilon}{{\boldsymbol \epsilon}}
\newcommand{\bftheta}{{\boldsymbol \theta}}
\newcommand{\bfxi}{{\boldsymbol \xi}}

\renewcommand{\grad}{{\boldsymbol \nabla\, }}

\draftfalse

\journalname{JGR: Machine Learning and Computation}
\addeditor{R1}
\addeditor{R2}
\addeditor{AU}
\newcommand{\RID}[1]{\textsuperscript{\scriptsize[\texttt{#1}]}}

\makeatletter
\renewcommand{\RID}[1]{%
  \if@keepnew
  \else
    \textsuperscript{\scriptsize[\texttt{#1}]}%
  \fi
}
\makeatother

\begin{document}

\pagestyle{fancy}
\fancyhead{} 
\renewcommand{\headrulewidth}{0pt}
\thispagestyle{fancy}

\title{Synthetic Geology: \\
Structural Geology Meets Deep Learning}

%
%




\authors{Simon Ghyselincks\affil{1},
Valeriia Okhmak\affil{3},
Stefano Zampini\affil{3},
George Turkiyyah\affil{3},
David Keyes\affil{3},
Eldad Haber\affil{2}}

\affiliation{1}{Department of Computer Science, University of British Columbia}
\affiliation{2}{Department of Earth, Ocean and Atmospheric Sciences, University of British Columbia}
\affiliation{3}{Division of Computer, Electrical and Mathematical Sciences and Engineering, King Abdullah University of Science and Technology}

\correspondingauthor{Simon Ghyselincks}{sghyseli@cs.ubc.ca}




\begin{keypoints}
\item Synthetic geology enables training of deep learning models for geophysics.
\item Generative flow matching reconstructs 3D geology from surface maps and sparse borehole data, giving multiple probable scenarios.
\item Supports geophysical inversions, probabilistic modeling, and resource exploration.
\end{keypoints}

%
%

%
%


\begin{abstract}
Reconstructing the structural geology and mineral composition of the first few kilometers of the Earth's subsurface from sparse or indirect surface observations remains a long-standing challenge with critical applications in mineral exploration, geohazard assessment, and geotechnical engineering. This inherently ill-posed problem is often addressed by classical geophysical inversion methods, which typically yield a single maximum-likelihood model that fails to capture the full range of plausible geology. The adoption of modern deep learning methods has been limited by the lack of large 3D training datasets. We address this gap with \textit{StructuralGeo}, a geological simulation engine that mimics eons of tectonic, magmatic, and sedimentary processes to generate a virtually limitless supply of realistic synthetic 3D lithological models. Using this dataset, we train both unconditional and conditional generative flow-matching models with a 3D attention U-\change{n}{N}et architecture. The resulting foundation model can reconstruct multiple plausible 3D scenarios from surface topography and sparse borehole data, depicting structures such as layers, faults, folds, and dikes. By sampling many reconstructions from the same observations, we introduce a probabilistic framework for estimating the size and extent of subsurface features. While the realism of the output is bounded by the fidelity of the training data to true geology, this combination of simulation and generative AI functions offers a flexible prior for probabilistic modeling, regional fine-tuning, and use as an AI-based regularizer in traditional geophysical inversion workflows.
\end{abstract}

\section*{Plain Language Summary}
Understanding and reconstructing the layers and structures of rock beneath the Earth's surface is an important but challenging task for geologists. It has practical uses in assessing geohazards such as earthquakes and landslides, and in guiding decisions for mineral and oil exploration. Most of our information for reconstruction comes from surface measurements, since direct observations are available only from narrow drill cores or mines. New and exciting artificial intelligence (AI) tools that can create realistic images or animations could be adapted \change{to}{for use in} geology, but they require large \change{swaths}{amounts} of training data\add{,} which we \change{do not have}{currently lack}. In this study, we develop \textit{StructuralGeo}, a geological computer simulation that reproduces geological processes over millions of years to create realistic examples of what lies hidden underground. The simulated data is used to train a generative AI that produces multiple 3D scenarios from existing surface maps and drill cores. The results can be used to enhance traditional techniques, for example\add{,} by estimating the possible size range of an underground rock body based on limited drilling samples.


%
%

\section{Introduction}

Structural geology focuses on the deformation and motion of rocks in the Earth's crust, describing three-dimensional distributions of rock formations and structures to understand the processes that shape the crust \cite{fossen2016structural}. Structural geologists use a variety of techniques and methods to investigate and interpret these features, including field mapping, structural analysis of rock outcrops, laboratory experiments, remote sensing, and geophysical imaging. The insights gained are essential for understanding the evolution of mountain ranges, the formation of mineral and energy resources, and \change{the assessment of}{assessing} geological hazards such as earthquakes and landslides, among other fundamental aspects of Earth science.

The construction of accurate structural geological models from collected data is a complex, ill-posed problem fraught with challenges \cite{mcclay2013mapping}. These models must reconcile geological observations obtained from what are often sparse and incomplete sources such as field data, rock samples, geophysical surveys, and remote sensing. Geological data is subject to uncertainty and interpretation, posing significant challenges in quantifying and addressing errors within the model \cite{wellmann2018ChapterOne3Da}. The interpretation of uncertain data is further exacerbated by the intricate and heterogeneous nature of geological structures, which unfold over diverse spatial and temporal scales ranging from microscopic mineral deformation to regional tectonic movements spanning millions of years \cite{karpatne2019MachineLearningGeosciencesa}. Integrating disparate datasets and models from multiple disciplines, including structural geology, geophysics, and geochemistry, represents an additional hurdle that demands interdisciplinary collaboration \cite{wellmann2010towards}.

Overcoming these challenges requires innovative technologies, advanced analytical methods, and interdisciplinary collaboration. However, the rewards of overcoming these obstacles are substantial, as accurate and reliable structural geological models are indispensable for understanding Earth's subsurface processes, predicting geological hazards, and managing natural resources. 
Traditionally, the construction of 3D structural geological models has been a manual, 
labor-intensive process, relying on expert interpretation, visualization, and iterative refinement of multiple data sources \cite{calcagno2008geological, houlding20123d}. Such manual workflows can take many hours and usually produce a single deterministic model, ignoring the underlying ill-posedness of the problem, \add[R2]{where multiple distinct subsurface configurations can fit the same set of observations.}\RID{R2-C2-2}

In this study, we take a step toward automating \change[R2]{inverse modeling}{subsurface reconstruction}\RID{R2-C2-3} of structural geology from \change[R2]{sparse or indirect observations (e.g., boreholes)}{sparse direct observations (e.g., boreholes) or indirect measurements}\RID{R2-C2-4} using generative artificial intelligence (GAI). Our goal is to develop a framework capable of parsing collected field data into coherent 3D geological reconstructions. Specifically, we aim to generate multiple plausible 3D models of lithology that reflect the range of geological scenarios consistent with a given set of observations. To achieve this, we \change[R2]{combine geological simulation with a deep generative learning framework.}{train a deep generative model using synthetic data produced by geological simulation.}\RID{R2-C3-2} 

First, in Sections \ref{subsec:Synthetic Dataset} and \ref{subsec:Sample Generation}, we develop \textit{StructuralGeo} \cite{StructuralGeo_v1}, a stochastic geology simulator that rapidly produces diverse 3D lithological volumes by composing representative tectonic, magmatic, and sedimentary events. Next, in Section \ref{subsec:synthgeoai}, we describe training a 3D flow-matching model \cite{albergoStochasticInterpolantsUnifying2023, lipman2023flow} fed by on-demand samples from \textit{StructuralGeo}. Flow matching regresses a continuous-time velocity field and is closely related to diffusion models \cite{yang2022diffusion}, providing a method of indirectly sampling from the distribution of simulated synthetic geology.

Once trained, the generative model, unlike the geology simulator, can be integrated with conditional data and applied to various geological tasks, described further in Section \ref{subsec:applications}. For example, it can be coupled with information obtained from drilling or surface measurement\add{s}, geostatistical interpolation, or geological segmentation that incorporates geological priors, improving model reliability and reducing ambiguity in subsurface interpretations. We demonstrate this capability in Section \ref{sec:Results} through experiments on the reconstruction of geological stratigraphy from borehole and surface lithology, illustrating the potential of generative models to advance structural geology research and practice.

\section{Methods}

\subsection{Synthetic Dataset}
\label{subsec:Synthetic Dataset}

Recent advances in generative diffusion models capable of strikingly realistic images \cite{yang2022diffusion} ha\change{s}{ve} been fueled by the availability of increasingly large, detailed training datasets such as Image\change{n}{N}et \cite{imagenet_cvpr09}. However, for 3D geological data, no comparable large-scale training sets exist, partly because we cannot directly observe high-resolution, three-dimensional underground rock formations. The primary obstacle is acquiring a comprehensive set of samples that reflects the wide variety of features and configurations that are present in the Earth's crust \cite{hadid2024WhenGeoscienceMeets}. To address this gap in information, we generate a synthetic dataset that captures the fundamental principles of structural geology\change{ that}{, which} have been amassed through research and field observation\add{s}. The ideal synthetic data for deep learning should simulate empirically drawn, independent, and identically distributed (i.i.d.) samples from the set of all 3D volumes in the Earth's crust. A deep generative model aims to learn a representation of the underlying probability distribution of geological structures from the samples, either implicitly or explicitly, in order to generate new samples that have never been seen before \cite{ruthotto2021IntroductionDeepGenerative}.

The difficulty of obtaining true observational data extends beyond geology, affecting other domains in scientific machine learning; fluid flow \cite{lienen2024zero} and biology \cite{goshisht2024machine} share similar challenges, for example. Moreover, \change{progressively more}{increasingly} complex large language models are \change{predicted}{expected} to \change{run out of sufficient}{encounter insufficient} training data in the future \cite{Jones2024}. In these cases, it has been suggested that numerical simulations or synthetic data \change{are}{should be} used to \change{either generate or augment the data that is needed to train a}{complement the available information for training} generative model\add{s}. Although such simulations can generate abundant, high-resolution data, they inevitably carry implicit biases that reflect the assumptions and simplifications of the underlying geological modeling. The realism of the simulated samples is ultimately bounded by those assumptions, which may introduce structural biases into any downstream inference. This limitation is not unique to synthetic data generation, as existing regularization and Bayesian inference methods likewise encode priors that shape and potentially bias the solution space \cite[Ch.~3.3]{somersallo}. Despite this challenge, such theoretical frameworks represent our best current knowledge about the particular field of interest, making them highly useful for \change[R2]{explaining}{interpreting}\RID{R2-C3-4} indirect measurements \add[R2]{by constraining the geology}\RID{R2-C3-4}. We view the framework presented ahead as a step toward\add{s} more specialized, \change[R2]{data-driven priors}{learned priors}\RID{R2-C3-5} that can evolve with \change[R2]{higher quality simulation}{increased realism in the choice of underlying simulation}\RID{R2-C3-5} and empirically evaluated simulation parameters.

In response to these constraints, we adopt a Geomodel-driven approach \cite{wellmann2018ChapterOne3Da}, bootstrapping an initial synthetic dataset through classical simulation methods based upon fundamental principles of how stratigraphic layers form over long time spans. We compile sequential, parameterized geological processes and workflows via mesh- and grid-based numerical techniques. This type of simulation is frequently used to explain observed geological formations. 

Current open-source 3D modeling efforts have been led by tools such as GemPy \cite{delaVargaOpenSourceModeling2019}, Noddy (\change{originally introduced}{introduced initially} in \cite{jessellNODDY1981} and more recently applied to large-scale synthetic dataset generation in \cite{jessellIntoNoddyverse2022}), and LoopStructural \cite{grose2021LoopStructural10Timeaware}, producing 3D geological volumes using a set of informed and constrained parameters to model limited underlying geological processes over time. While these tools are invaluable in their respective scopes of use, they do not natively support large-scale randomization. For example, \citeA{jessellIntoNoddyverse2022} have suggested that \add{incorporating} more parameters, more events, linked events, and more models would \change{improve}{enhance} their existing efforts. We thus adopt a computational method suggested by \citeA{cockett2016visible}, where geological processes are represented as mathematical transformations of a 3D mesh. By combining multiple transformations, one can generate a wide range of realistic geological models.
In fact, such models are used to train geologists \cite{cockett2016visible}, suggesting their suitability as a synthetic dataset for training machine-learning models. Additionally, randomizing and reordering these processes enables the generation of a virtually limitless number of 3D volumes depicting complex and varied geology. We implement this randomized workflow in the open‑source Python package \emph{StructuralGeo}.

\subsection{Sample Generation}
\label{subsec:Sample Generation}
\add[R2]{Following} \citeA{wellmann2018ChapterOne3Da},\add[R2]{ we define a Geomodel (or structural geological model) as a representation of geometric elements such as rock unit boundaries and faults on a scale of meters to kilometers. Our }\RID{R2-C2-1}
\remove[R2]{Geological model training data samples, referred to as }Geomodels take the form of 3D tensors denoted as $\bfm(\bfx):\mathbb{R}^3 \rightarrow {\mathbb N}, \ \ \bfx = [\bfx_1, \bfx_2, \bfx_3]$ and ${\mathbb N} = \{1, \ldots, N\}$. Each element of the tensor $\bfx$ corresponds to a voxel in a 3D mesh grid, with an integer label (from $1,\ldots, N$) indicating its assigned rock-strata category. The data is composed of discrete categorical values, which are representative of the labeling and segmentation that a geologist would apply to collected field data\add[R2]{, distinct from continuous physical property fields}\RID{R2-C2-5}.

We define two classes of processes acting on $\bfx$ :
\begin{enumerate}
    \item {\em Transformations} $\mathcal{T}: \mathbb{R}^3 \rightarrow \mathbb{R}^3$ which transport or displace the geological medium, and
    \item {\em Depositions} $\mathcal{D}: \mathbb{R}^3  \rightarrow \mathbb{N}$ which modify the composition.
\end{enumerate}

\subsubsection{Transformations}
Transformations $(\mathcal{T})$ are general operations that deform, rotate, or otherwise displace the geological medium according to parameters~$\bftheta$.
Let $\bfu(\bfx; \bftheta) \colon \mathbb{R}^3 \to \mathbb{R}^3$ be a vector-valued function dependent on~$\bftheta$. A single transformation ${\cal T}_{\bftheta}$ is then expressed as a coordinate mapping, that is
\begin{linenomath*}
\begin{eqnarray}
    \label{eq:geoproc}
    \mathcal{T}_{\bftheta}\bigl(\bfm(\bfx)\bigr) 
    := \bfm\bigl(\bfx + \bfu(\bfx; \bftheta)\bigr).
\end{eqnarray}
\end{linenomath*}
Transformations describe many geological processes ranging from folding to shearing and are combined to create complex deformations. However, they are insufficient if we consider the need to introduce or remove geological materials from the Geomodel domain, for which we use deposition events.

\subsubsection{Depositions}
Depositions $(\mathcal{D})$ add or remove material into the model domain through sedimentation, erosion, or intrusion events; for example, the transport of hot fluids from the interior of the Earth through a fissure.
In a deposition, we envision some material (new rock) that replaces existing rock by pushing it aside, eroding it away, or accumulating above it. The extent and nature of a deposition event are controlled by a set of parameters~$\bftheta$.  A simple mathematical model uses a window function $\bfw(\bfx, \bftheta)$ that has \change{the}{a} value \add{of}~$1$ where the deposition occurs and~$0$ otherwise. This function is generated by first considering a canonical window
$$\bfw_0(\bfx) := \prod_{i=1}^3(H(\bfx_i+1) - H(\bfx_i-1)), $$
where $H$ is the Heaviside function. We then map $\bfw_0(\bfx)$ using a transformation $\mathcal{T}_{\bftheta}$ to allow scaling, rotations, translations, and bending, that is
\begin{linenomath*}
\begin{eqnarray}
    \label{eq:wind}
    \bfw(\bfx, \bftheta) = {\cal T}_{\bftheta}(\bfw_0(\bfx)).
\end{eqnarray}
\end{linenomath*}
Given a Geomodel $\bfm(\bfx)$ and a category label \(a\), the deposition process is defined by
\begin{linenomath*}
\begin{eqnarray}
    \label{eq:deposition}
    \mathcal{D}_{\bftheta}\bigl(\bfm, a\bigr) 
    = \bigl(1-\bfw(\bfx,\bftheta)\bigr) \,\bfm(\bfx) 
    + a\,\bfw(\bfx,\bftheta).
\end{eqnarray}
\end{linenomath*}
Hence, \(\bfm(\bfx)\) is overwritten by the new category \(a\) wherever \(\bfw(\bfx,\bftheta) = 1\).

\subsubsection{Geological Histories}
\begin{figure}[t]
    \centering

    \includegraphics[width=\linewidth]{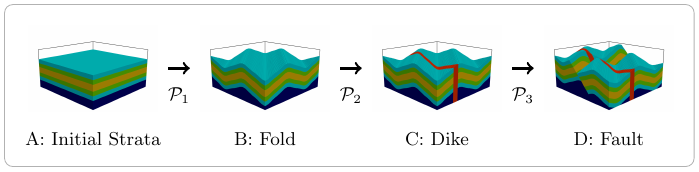}
    \caption{%
    Example of geological history illustrating the application of transformations and depositions. Panels:
    (A) initial strata $\bfm_0(\bfx)$,
    (B) fold $\mathcal{T}_{\text{fold}}$,
    (C) dike $\mathcal{D}_{\text{dike}}$,
    and (D) fault $\mathcal{T}_{\text{fault}}$.
    Processes $\mathcal{P}_i$ are sequenced randomly from a set ${\cal P}_j, j=1,\ldots,k$ using a Markov chain, leading to varied outcomes.
    }
    \label{fig:geoprocess}
\end{figure}

A complete Geomodel is constructed by combining both transformation $(\mathcal{T})$ and depositions $(\mathcal{D})$ to simulate geological evolution over discrete time steps. The initial state of the Geomodel is described by a mapping $\bfm_0(\bfx)$, which depends only on the vertical axis $\bfx_3$, thus forming sequential horizontal layers. Next, multiple transformations and depositions (e.g., folding, dike formation, shearing), each with its parameters $\bftheta$, are sequenced together to form a more complex history.
Let ${\cal P}^{i}$ represent a generic geological process (either $(\mathcal{T})$ or $(\mathcal{D})$). The complete history is simply a composition of many such processes, that is
\begin{linenomath*}
\begin{eqnarray}
    \label{eq:history}
  \bfm(\bfx) =   {\cal H}(\bfm_0) =  \mathcal{P}^{n} \circ \cdots\circ \mathcal{P}^{i+1} \circ \mathcal{P}^{i}\cdots
  \circ \mathcal{P}^{1} \bigl(\bfm_0(\bfx)\bigr).
\end{eqnarray}
\end{linenomath*}
To numerically evaluate the model $\bfm(\bfx)$ under transformations and depositions, we use backward interpolation, common in image transformation problems \cite{Modersitzki2004}. Figure~\ref{fig:geoprocess} illustrates one example of such a history. 

\begin{figure}[t]
\centering
 \includegraphics[width=.8\linewidth]{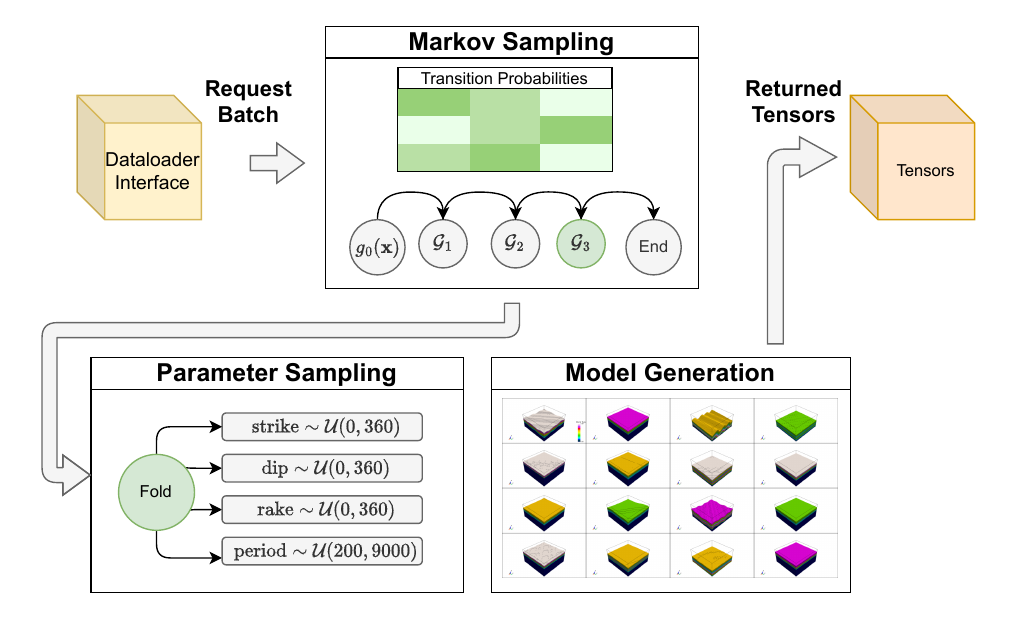}
\caption{%
Pipeline for random geological model generation.
A batch request triggers a Markov chain sampler which selects and sequences geological processes (e.g., folding, deposition) from a transition matrix; parameters such as amplitude and wavelength are drawn from tuned random variables; the sequences are applied to produce batches of 3D tensors.}
\label{fig:dataloader-pipeline}
\end{figure}

Although the Geomodel \eqref{eq:history} is deterministic once the sequence and parameters are fixed, we introduce variability using a Markov chain to sequence new histories, drawing from the set of all possible geological processes $\{{\cal P}^j\}, j=1,\ldots,k$. After each process, the next is chosen with some probability 
\begin{linenomath*}
\begin{eqnarray}
    \label{eq:markov}
    \pi_j = \pi({\cal P}^j|{\cal P}^{j-1}),
\end{eqnarray}
\end{linenomath*}
conditioned on the previous event. This yields a random geological history for the computational sequence in \eqref{eq:history}. Termination occurs when a special end-of-process token is drawn; hence, models emerge as random computational sequences of variable length.
 Because each process has its randomized parameters, repeated runs produce diverse realizations with high variability in both the ordering of events and overall geological age. Figure~\ref{fig:dataloader-pipeline} provides an overview of the generation pipeline.

\subsubsection{Limitations}

The above simulation can generate a virtually infinite number of geological models with variable complexity and at any specified resolution, which can be used as data for a generative model. Nonetheless, just like any other simulation, it depends on a number of hyperparameters. Every geological process depends on a set of parameters that are sampled as random variables and must be tuned such that the outcome is ``realistic" as visually determined by geologists. In addition, the transition probabilities $\pi_j$ between processes must be chosen. As part of \add{the} development \add{process}, representative samples were shown to domain experts to verify the large-scale geometries produced by the simulator\change{, though}{; however,} feedback has been strictly qualitative and informal. \change{However}{Despite this}, it is important to realize that such models represent what experts consider to be unbiased geology. Since there are no direct observations on the geology at the (3D) resolution and scale that we need, it is virtually impossible to know if such models introduce implicit bias. Nonetheless, such models are extremely useful in generating what experts believe is a realistic Earth model.

While multiscale and multifractal variability are well established characteristics of natural geological and geophysical fields (e.g., \cite{Dolan98, li2003scale, gagnon2003, lovejoy2007}), the present simulations \change[R2]{target a large scale regime and}{are configured to resolve large-scale structures}\RID{R2-C4-1} with discrete lithological categories. The simulation dataset is configured to capture characteristics in the $30-60\,\text{m}^3$ resolution range, emphasizing the large-scale geometry and spatial arrangement of structural domains rather than fine-scale heterogeneity. The segmentation of geology into discrete rock types serves to \change[R2]{remove variation in favor of large-scale classification}{intentionally pool fine-scale heterogeneity into broader structural units}.\RID{R2-C4-1} The current implementation of \textit{StructuralGeo} is constrained to applications where this type of variability is not desired or required.

\subsection{Synthetic Geology Using Generative AI}
\label{subsec:synthgeoai}
We define the simulator’s output distribution of Geomodels over all possible 3D volumes as $\pi_\bfm$, and aim to train a GAI model that can approximate \change[R2]{and draw new samples from this distribution}{$\pi_\bfm$ to perform unconditional sampling and conditional probabilistic reconstruction}\RID{R2-C3-3} without relying on the original simulation engine. The simulator provides the training data, and the GAI learns to produce new examples from $\pi_\bfm$ in a probabilistic manner, analogous to training an image generator on a large dataset of photos. However, simulator outputs have the advantage of being computed and tuned on demand, providing endless variation with a low memory storage footprint compared to fixed datasets.

Let $\bfm(\bfx)$ be a geological model generated by the simulation process using a sampled geological history ${\cal H}(\bfm_0)$, see \eqref{eq:history}. Given a method to generate virtually unlimited samples $\bfm(\bfx)\!\sim\!\pi_\bfm$, we now have the tools to train a foundation model that indirectly learns the underlying probability characteristics of any given Geomodel, $\pi_{\bfm}(\bfm)$. For example by learning its associated score function $s_\bfm := -\grad_\bfm \log (\pi_\bfm(\bfm))$ \cite{song2019generative}, or, as in our case, through the related flow-matching objective described in Section~\ref{subsubsec:flowmatching}. Note that $\bfm$ represents discrete categorical data, so $\pi_\bfm$ is not directly differentiable with respect to $\bfm$. We address this by adopting a continuous embedding and decoding scheme that enables differentiation in the model’s latent space.

A natural question arises: what are the key advantages and practical applications of learning the distribution $\pi_\bfm$ associated with the synthetic generative process described earlier? That is, how does the ability of a foundation model to estimate $\pi_\bfm$ enhance our understanding or extend our capabilities, given that we already have a method of generating such samples?

 While the original simulation process can generate diverse Geomodels, it is \change{hard}{challenging}, and in some cases virtually impossible, to use the simulation for any \change{other application besides}{application other than} the forward process of generating a random 3D geological volume from a parameterized history ${\cal H}(\bfm_0)$. In contrast, foundation models support a broader range of applications\change{ that vary}{,} from generation to conditional generation or inversion, enabling the incorporation of other data or constraints based on conditional distributions \cite{chung2022come}. 
 
For instance, while the simulation process requires a structured history of geological events, a foundation model can generate realistic samples conditioned directly on partial observations in the spatial domain $\bfx$, such as borehole data or surface measurements. Since our ultimate goal is to \change{fuse}{integrate} multiple sources of geoscientific data, we seek to replace the rigid simulation process with a flexible, trainable foundation model.

One approach to learning a probability density function $\pi_\bfm$ is the denoising or diffusion model, now commonly used for image and movie generation. While diffusion methods are often applied to continuous-valued data, the generation of discrete-valued data has only more recently been addressed by \citeA{meng2022concrete} through the theory first proposed by \citeA{hyvarinen2007some}. 
A more general \change{way to approach the problem of}{approach to} matching distributions that does not require the score is flow matching.

\subsubsection{Flow Matching}
\label{subsubsec:flowmatching}
Flow matching, also know\add{n} as stochastic interpolation, is a powerful technique to learn a continuous normalizing flow between two probability distributions \cite{lipman2023flow,albergoStochasticInterpolantsUnifying2023}. Rather than attempt to directly estimate a complex probability distribution such as $\pi_\bfm$, the model learns a velocity field that transports samples from a simpler source distribution $\pi_0$ at $t=0$ to our target distribution $\pi_\bfm$ at $t\!=\!1$.

Let $t\in[0,1]$ be the time parameter, let $\bfm^{(e)} = \text{enc}(\bfm(\bfx))$ denote the continuous latent embedding of $\bfm(\bfx) \sim \pi_\bfm$ (see Equation~\eqref{eq:full_emb}), and let $\bfz \sim N(0, \bfI)$ be a Gaussian random variable drawn from the source distribution. Then the stochastic interpolant $\bfm_t^{(e)}$ between the start and target distributions is defined as
\begin{linenomath*}
\begin{eqnarray}
    \label{eq:Gt}
    \bfm_t^{(e)} = t \bfm^{(e)} + (1-t)\bfz,
\end{eqnarray}
\end{linenomath*}
and the velocity between the source and target is
\begin{linenomath*}
\begin{eqnarray}
    \label{eq:flow_vel}
    \bfv_t = \frac{d}{dt}\bfm_t^{(e)} = \bfm^{(e)} - \bfz.
\end{eqnarray}
\end{linenomath*}
Flow matching seeks to learn the unique minimizer over all $t, \mathbf z, \mathbf m^{(e)}$ of the quadratic objective
\begin{linenomath*}
\begin{eqnarray}
    \label{eq:stopt}
    \min_{\boldsymbol\xi} \frac{1}{2}\, \mathbb E_{t, \mathbf z, \mathbf m^{(e)}} 
    \left\| \mathbf v(\mathbf m_t^{(e)}, t; \boldsymbol\xi) - \mathbf v_t \right\|_2^2 .
\end{eqnarray}
\end{linenomath*}
In practice the true minimizer is approximated by a deep neural network
\begin{linenomath*}
\begin{eqnarray}
    \label{eq:network}
    \hat{\mathbf v}(\mathbf m_t^{(e)}, t; \boldsymbol\xi),
\end{eqnarray}
\end{linenomath*}
parameterized by $\boldsymbol\xi$ and trained using stochastic gradient descent.

At inference time, samples are drawn from $\pi_0$ and integrated through the learned velocity field $\hat \bfv$ using either an ordinary differential equation (ODE) or stochastic differential equation (SDE) solver to get novel samples from the embedded space of $\pi_\bfm$. In our method\add{,} we use the deterministic ODE method with $\pi_0 = N(0, \bfI)$, as it is easier to work with and does not appear to compromise sample quality.

\subsubsection{Embedding Discrete Categorical Data}

To adapt flow matching to discrete categorical geology models, we embed the model into a continuous vector space without imposing any ordering on the $N$ categories. 
A Geomodel is a discrete field $\bfm:\mathbb{R}^3 \rightarrow \{1,\ldots, N\},$
assigning to each location $\bfx$ a categorical lithology label.

We map each category to a normalized embedding vector through
\begin{linenomath*}
\begin{eqnarray}
    \label{eq:emb_function}
\text{enc}(i) = \frac{e_i - \frac{1}{N}\mathbf{1}}{1 - \frac{1}{N}},
\end{eqnarray}
\end{linenomath*}
where $e_i$ is the $i$-th standard basis vector and $\mathbf{1}$ is the length-$N$ vector of ones.  
The embedded model is obtained pointwise as
\begin{linenomath*}
\begin{eqnarray}
    \label{eq:full_emb}
\bfm^{(e)}(\bfx) = \text{enc}(\bfm(\bfx)).
\end{eqnarray}
\end{linenomath*}
Applying $\text{enc(.)}$ across the entire domain yields a continuous field 
$\bfm^{(e)} \in \mathbb{R}^{[N,X,Y,Z]}$ on a mesh of size $[X,Y,Z]$. 
This continuous representation $\bfm^{(e)}$ resides in a high\add{-}dimensional space\add{,} for which we can compute gradients with respect to the embedded model parameters $\bfm^{(e)}$ for flow matching, \change{see}{as shown in} Equation~\eqref{eq:stopt}.

 Because we do not constrain the embedded vectors to the probability simplex, the flow field is free to traverse an unconstrained continuous space. It is then the task of the model to learn the equivalent probability distribution $\pi_{\bfm^{(e)}}$ in the embedded space, from which we can decode back to the original categorical space of $\bfm$. 
 
 To decode a voxel $\bfm^{(e)}(\bfx)\in \mathbb{R}^N$ back to a discrete category $\bfm(\bfx)=i \in \mathbb{N}$, we select the embedding vector over all $j \in \mathbb{N}$ with the largest cosine similarity:
 \begin{linenomath*}
\begin{eqnarray}
    \label{eq:decoder}
\text{dec}(\bfm^{(e)}(\bfx)) = \arg\max_{j} \langle \bfm^{(e)}(\bfx), \text{enc}(j) \rangle.
\end{eqnarray}
\end{linenomath*}
Since the embedding vectors are unit-norm, this is equivalent to nearest-neighbor or shortest \change{e}{E}uclidean distance.

\subsection{Applications to Inverse Problems}
\label{subsec:applications}

A trained model $\hat \bfv(\bfm_t^{(e)}, t; \bfxi)$ indirectly learns the distribution $\pi_\bfm$ through the latent embedding space, allowing us to draw or infer new geological models that are similar to the training set. Generating arbitrary samples from $\pi_\bfm$ is an example of unconditional generation where the flow process is not guided or conditioned by any constraints. To draw a new sample, we integrate the learned ordinary differential equation (ODE) 
\begin{linenomath*}
\begin{eqnarray}
\label{eq:flow}
{\frac {d\bfm}{dt}} = \hat \bfv(\bfm_t^{(e)}, t; \bfxi) \quad \bfm_0^{(e)} \sim N(0,\bfI)\ \quad t\in [0,1],
\end{eqnarray}
\end{linenomath*}
starting from pure Gaussian noise at $t=0$, using a numerical solver such as Runge-Kutta or adaptive Heun's method.

While \add{the }unconditional generation of models is helpful in many applications where models are needed for \change{the simulation of}{simulating} other data (see, e.g., \citeA{wellmann2015PynoddyExperimentalPlatform}), a more important application is the integration of measured data. 
Suppose we have noisy observations of the form
\begin{linenomath*}
\begin{eqnarray}
    \label{eq:data}
    \bfd = \bfA \bfm + \bfepsilon, 
\end{eqnarray}
\end{linenomath*}
where $\bfA$ is the forward mapping \add[R2]{such as a sampling mask or physical operator}\RID{R2-C4-2}, $\bfepsilon \sim N(0, \sigma^2 \bfI)$ is Gaussian noise, and $\bfd$ is an indirect observation of $\bfm$.

Equation \eqref{eq:data} is an algebraic constraint that is added to the original ODE from Equation \eqref{eq:flow}, turning it into a Differential Algebraic Equation (DAE) \cite{AscherPetzoldODEs}, a formulation that recently has been explored in deep learning contexts \cite{Boesen2025DAEs}.
A straightforward approach is adding a term to the ODE that attracts the solution towards the measured data, obtaining the system
\begin{linenomath*}
\begin{eqnarray}
    \label{eq:flow_cons}
{\frac {d\bfm}{dt}} = \bfv(\bfm_t, t; \bfxi) - \mu \bfA^{\top}(\bfA \bfm - \bfd) \quad \bfm_0 \sim N(0,\bfI)\ \quad t\in [0,1].
\end{eqnarray}
\end{linenomath*}
The parameter $\mu$ balances between data fit and fit to the prior model and is tuned typically by cross-validation. 

\subsubsection{Conditional Flow Matching}
For a general application of conditional data to the generation process, we add an additional embedding of conditional data $\bfd$ to the machine learning model and a modification to the loss function using a method adapted from recent work by \citeA{ahamedhaber2024}. This conditional flow matching (CFM) model \cite{lipman2023flow} learns the conditional velocity field $\hat \bfv(\bfm_t, \bfd, t; \bfxi)$ which represents the posterior distribution $\pi_{\bfm|\bfd}$. Conditioning enters through the \change[R2]{embedded adjoint $\mathbf{A}^\top \mathbf{d}$}{embedding of the adjoint-data product $(\mathbf{A}^\top \mathbf{d})^{(e)}$}\RID{R2-C4-3}, permitting the flow field to adapt according to both the latent representation $\bfm_t$ and the observed data $\bfd$. From Equation \eqref{eq:flow} the predictor
\begin{linenomath*}
\begin{eqnarray}
\label{eq:dhat}
\hat \bfd = \bfA \hat \bfm = \bfA \left(\bfm_t + (1-t)\cdot \hat \bfv(\bfm_t, \bfd, t; \bfxi)\right)
\end{eqnarray}
\end{linenomath*}
provides a differentiable estimate of the observable response. Since this has high variance for $t \rightarrow 0$ but must match the data as $t \rightarrow 1$, we introduce a time-weighted conditional loss,
\begin{linenomath*}
\begin{eqnarray}
\label{eq:cond_loss}
\mathcal{L}_{\text{cond}} = \left\|\hat \bfv(\bfm_{t}, \bfd, t; \bfxi) - \bfv_t\right\|_2^2+\lambda \cdot t \cdot \left\|\hat \bfd - \bfd \right\|_2^2.
\end{eqnarray}
\end{linenomath*}
The multiplying factor of $t$ ensures that the data consistency is emphasized only near the end of the flow trajectory, preventing instabilities in training. This extends CFM to inverse problems by jointly learning both the conditional velocity field and the reconstruction implicit in $\hat \bfd$.

Ultimately, the goal is to learn to sample from the posterior distribution $\pi_{\bfm|\bfd}$ to produce \change{many different}{multiple} candidate geological models consistent \change{in}{with} $\bfd$.

\subsection{Network Architecture}
\label{subsec:architecture}

So far\add{,} we have presented a method of reconstructing samples from both $\pi_\bfm$ and $\pi_{\bfm|\bfd}$. Now we describe our implementation of a deep learning network for modeling the conditional estimator $\hat \bfv(\bfm_t^{(e)}, \bfd, t; \bfxi)$, noting that the unconditional generation operates with a similar architecture but \change[R2]{without conditional embeddings}{omits the conditional inputs $(\mathbf{A}^\top \mathbf{d})^{(e)}$}\RID{R2-C5-1}. To learn $\hat \bfv$, we use a magnitude preserving attention 3D U-Net \cite{vaswaniAttentionAllYou2017, ronneberger2015u, Karras2023AnalyzingAI}. The model combines 3D convolutional skip connections and attention layers to capture multiscale relationships, both locally and at long range. Self-attention layers are particularly beneficial for geological constructions where long-range stratigraphic continuity and spatial correlation cannot be captured solely by convolutional operations.

The model requires a scheme for embedding and injecting both $\bfA^{\top}\bfd$ and $t$. At each resolution scale of the U-Net, a module \add[R1]{$\bfg(.)$}\RID{R1-C2} composed of convolution and activation layers provides feature extraction, and rescaling of $(\bfA^{\top}\bfd)^{(e)}$. We perform time embedding with either sinusoidal \cite{vaswaniAttentionAllYou2017}, random Fourier \cite{rahimirandfour}, or learnable Fourier features \cite{tancikFourierFeaturesLet2020}\add{,} which are then passed through a multi-layer perceptron (MLP). Finally\add{,} a mixing module concatenates feature tensor $\bfx_n$ with the embedded adjoint \change[R2]{$\bfA^{\top}\bfd$}{$(\bfA^{\top}\bfd)^{(e)}$}\change[R1]{, then applies a feature-wise linear modulation (FiLM) through a residual connection}{. The combined features are modulated by the time embedding using a feature-wise affine transform (FiLM)} \cite{perez2018film} \add[R1]{, then processed through a convolutional block $\mathcal{F}$ and added back to $\bfx_n$:}
\begin{linenomath*}
\begin{equation}
    \label{eq:film_mixing}
    \mathbf{x}_{n+1} = \mathbf{x}_n + \mathcal{F}\left( 
    \underbrace{(\mathbf{1} + \boldsymbol{\gamma}_t)}_{\text{time scale}} \odot 
    [\mathbf{x}_n, \underbrace{\bfg((\bfA^{\top}\bfd)^{(e)})}_{\substack{\text{scaled cond.} \\ \text{features}}}] 
    + \underbrace{\boldsymbol{\beta}_t}_{\substack{\text{time} \\ \text{shift}}} 
    \right),
\end{equation}
\end{linenomath*}
 \add[R1]{where $[\cdot, \cdot]$ denotes channel-wise concatenation, $\odot$ is the Hadamard product, and $(\boldsymbol{\gamma}_t, \boldsymbol{\beta}_t)$ are affine parameters from the time embedding.}\RID{R1-C2}

\change[R1]{A sample $\bfm$ is drawn from the simulator,}{Each training iteration begins by drawing a random sample $\bfm$ from the simulator,}\RID{R1-C3} together with optional conditional observations from Equation~\eqref{eq:data} and their adjoint $\bfA^\top \bfd$. Both tensors are passed through an encoder into a continuous space for flow matching. \change[R1]{A stochastically sampled pair $\bfz, t$ is drawn}{Next, a noise sample $\bfz$ and time $t$ are drawn,}\RID{R1-C3} and Equation~\eqref{eq:Gt} is used to generate the interpolated state $\bfm_t^{(e)}$ which serves as the training input for the \add[R1]{time-dependent}\RID{R1-C2} velocity network $\hat \bfv$, implemented as a 3D U-Net. The time embedding conditions the network on $t$, \add[R1]{globally modulating the network layers and outputs,}\RID{R1-C2} while for CFM the conditional observations $\bfd$ are included via concatenated embeddings at each U-Net scale. Attention layers broadly distribute information spatially, while the skip connections transmit across resolutions.

The true velocity target $\mathbf{v}_t$ is computed \remove[R1]{using Equation (7)} from the \change[R1]{withheld pair $(\mathbf{m}^{(e)}, \mathbf{z})$}{underlying components $(\mathbf{m}^{(e)}, \mathbf{z})$ of the interpolated state} \add[R1]{using Equation}~\eqref{eq:flow_vel}\RID{R1-C3}, and Equation~\eqref{eq:dhat} yields the reconstructed observations $\hat{\mathbf{d}}$. The predicted velocity $\hat{\mathbf{v}}_t^{(e)}$ is compared with $\mathbf{v}_t$ to form the conditional loss $\mathcal{L}_{\text{cond}}$, which backpropagates to update both the learnable time embeddings and the U-Net parameters.

\begin{figure}[t]
    \centering
    \includegraphics[width=1.0\textwidth]{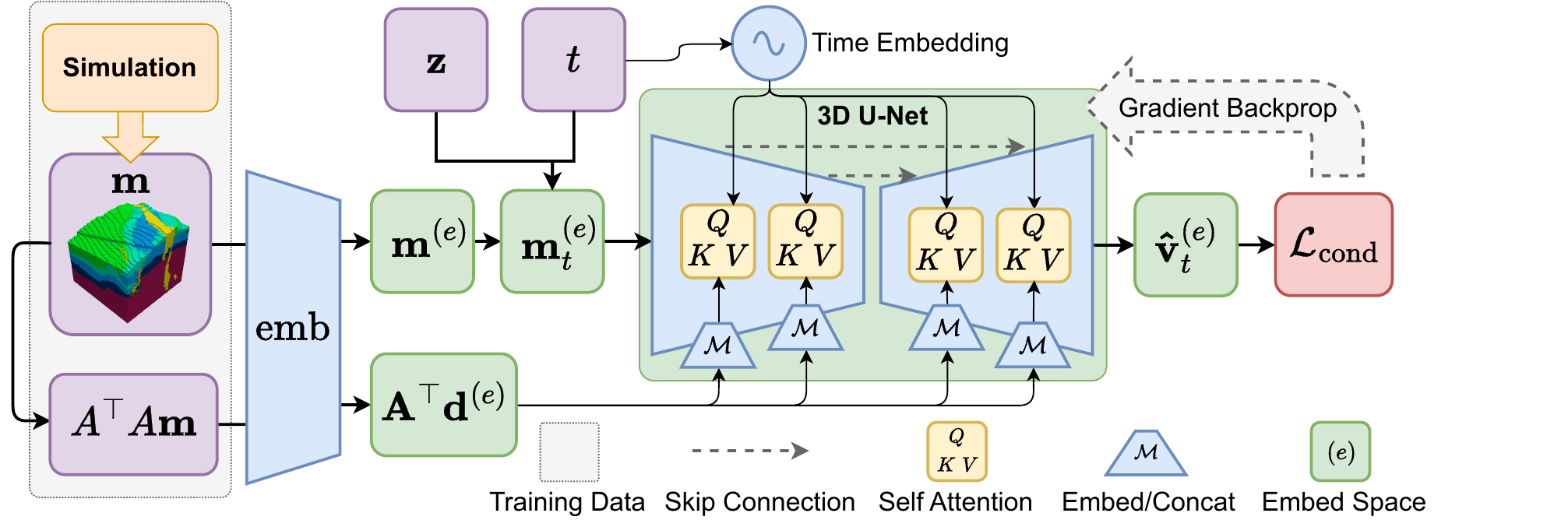} 
    \caption{Overview of our conditional flow-matching architecture. 
The diagram is inspired by the Latent Diffusion Models paper (Rombach et al., 2022).}
    \label{fig:arch}
\end{figure}

For training stability, following the work of \citeA{albergoStochasticInterpolantsUnifying2023}, we sample a reduced time range of $t \in [0.0001, 0.9999]$ and use gradient clipping, \change[R2]{with an exponential moving average (EMA) of the weights at inference}{maintaining an exponential moving average (EMA) of the weights for use at inference}\RID{R2-C5-3} \cite{tarvainenema2017}. We found that the learnable Fourier features provided the best stability in training. Our architecture and hyper\remove{-}parameter selection \change{was}{were} determined first through experiments with 2D U-Nets using CIFAR-10 for image generation, then with 3D U-Nets using lower resolution $16\times16\times16$ simulation data, \change{then}{and} finally scaling to $64\times64\times64$ simulation data. Architectural details, training procedures, and model hyperparameters are documented in the code base \cite{FlowTrain_v1}.

\section{Results}
\label{sec:Results}
\subsection{Unconditional Generation of Earth Models}

\begin{figure}[t]
\centering
\includegraphics[width=0.98\textwidth]{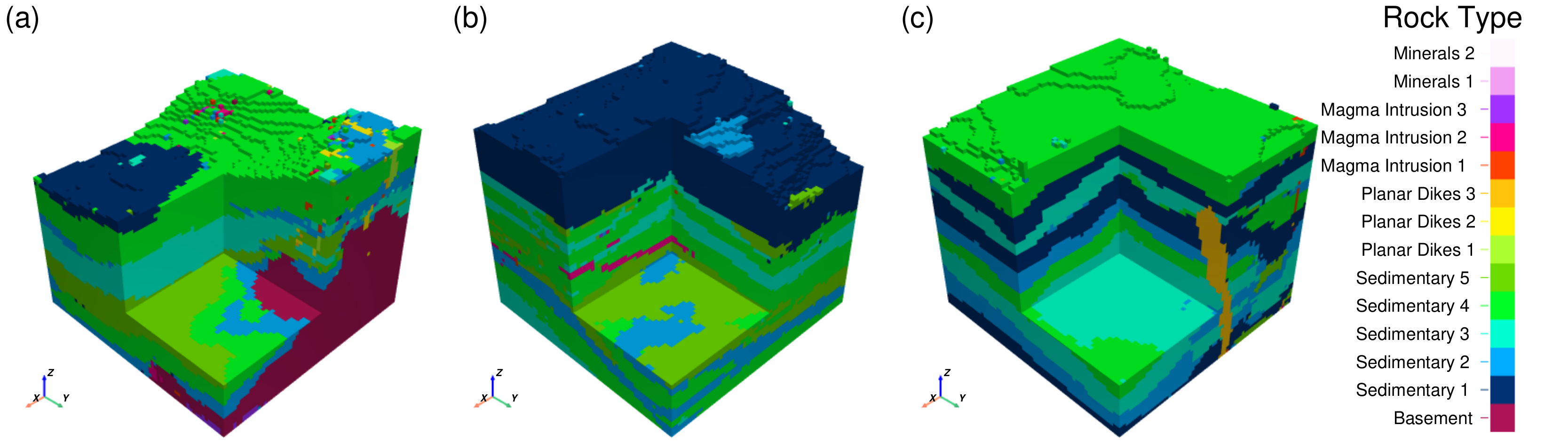}
\caption{\textbf{Three examples of unconditional earth models.} Panels (a–c)  $64 \times 64 \times 64$ resolution generative samples from $\pi_{\bfm}$ using flow matching. Each model represents a $3.84\, \text{km} \, \times 3.84\, \text{km} \, \times 3.84\, \text{km}$ volume of the Earth's crust. Section cutaways into the generated model are added to show portions of the interior.}
\label{fig:unconditioned-samples}
\end{figure}

In our first set of experiments, we show that a flow matching model trained with suitable synthetic data in a continuously embedded space can generate realistic earth models. Unconditional generation is carried out using the learned velocity flow field and a numerical ODE solver \cite{torchdiffeq}. After integration, the resulting sample is decoded into the categorical space, as formulated in \eqref{eq:decoder}, giving 3D categorical samples as shown in Figure~\ref{fig:unconditioned-samples}.

These models are diverse, encompassing variations in topography, layer thickness and geometry, as well as d\change{y}{i}kes and intrusions. Such samples can support data generation for applications like seismic simulation \cite{woollam2021seisbenchtoolboxmachine}, geological segmentation and mapping \cite{julka2023knowledgedistillationsegmentsam}, and classification \cite{Zhang2018GeoSeg}.

\subsection{Conditional Generation of Earth Models}

In our next experiment, we demonstrate how conditional generation can be \change{used for}{applied to} a common \change{but}{yet} important task: geological interpolation of lithology from borehole data. The procedure entails \change{constructing}{the construction of} 3D subsurface models from sparse, irregularly sampled borehole data. This task bridges observed lithological measurements (e.g., sand, clay, shale) with geostatistical or simulation-based techniques to predict the unobserved subsurface composition. Such a problem is commonly solved to create initial models for reservoir characterization, groundwater flow analysis, or contaminant transport studies.

The standard approach to this problem is kriging \cite{stein1999interpolation} or, more recently, interval kriging \cite{Song2024Kriging}, which estimates lithological probabilities between boreholes by accounting for vertical and horizontal variability in rock types. 
However, these methods can struggle with sparse data (\add{i.e., }few boreholes) and \change{have the tendency}{tend} to smooth interfaces when applied to categorical data. 
We demonstrate an alternative reconstruction method using a conditional flow matching model to interpolate between sparse boreholes. Because the technique is inherently generative, we can produce \change{many different}{multiple} models that are consistent with \add{both }the observed data and \remove{also with }the underlying probability density function of the synthetic training dataset.

\begin{figure}[t]
  \centering
    \includegraphics[width=\linewidth]{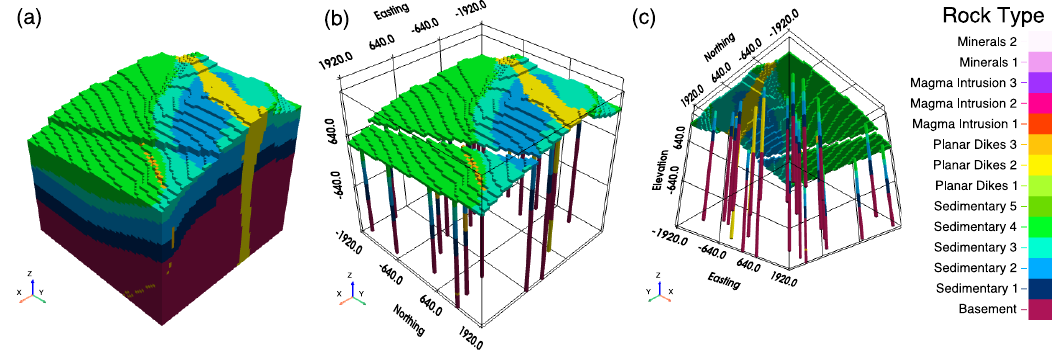}
    \caption{(a) A 3D Geomodel unseen during training is used to produce sparse conditional data for reconstruction in (b) and (c) consisting of 25 randomly placed vertical boreholes one single $(60 \times 60 \times 60)\text{m}^3$ voxel in diameter.}
    \label{fig:model3d+data}
\end{figure}

To guide this example, we extract the observable surface data and sparse boreholes from a Geomodel that was never seen during training and use the original model as a ground truth benchmark. Shown in Figure~\ref{fig:model3d+data} are the original mode\add{l}, the surface data, and 25 boreholes for the experiments that follow.

\begin{figure}[h]
    \includegraphics[width=\linewidth]{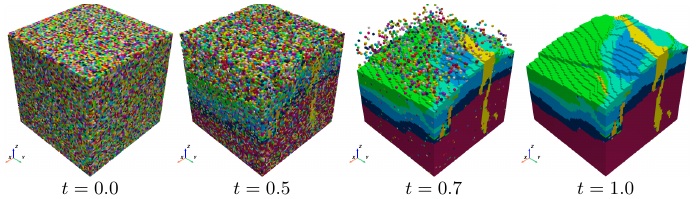}
    \caption{Generative trajectory from Gaussian noise to geology conditioned on sparse data. Decoded categorical models at intermediary times $t\in[0,1]$ illustrate the continuous flow toward $\pi_{\bfm|\bfd}$.}
    \label{fig:recfrombhtime}
\end{figure}

The surface data and boreholes are used to guide the flow process using a set of different starting seeds consisting of Gaussian noise drawn from a normal distribution, shown in Figure~\ref{fig:recfrombhtime}.

\begin{figure}[t]
    \centering
    \includegraphics[width=.98\textwidth]{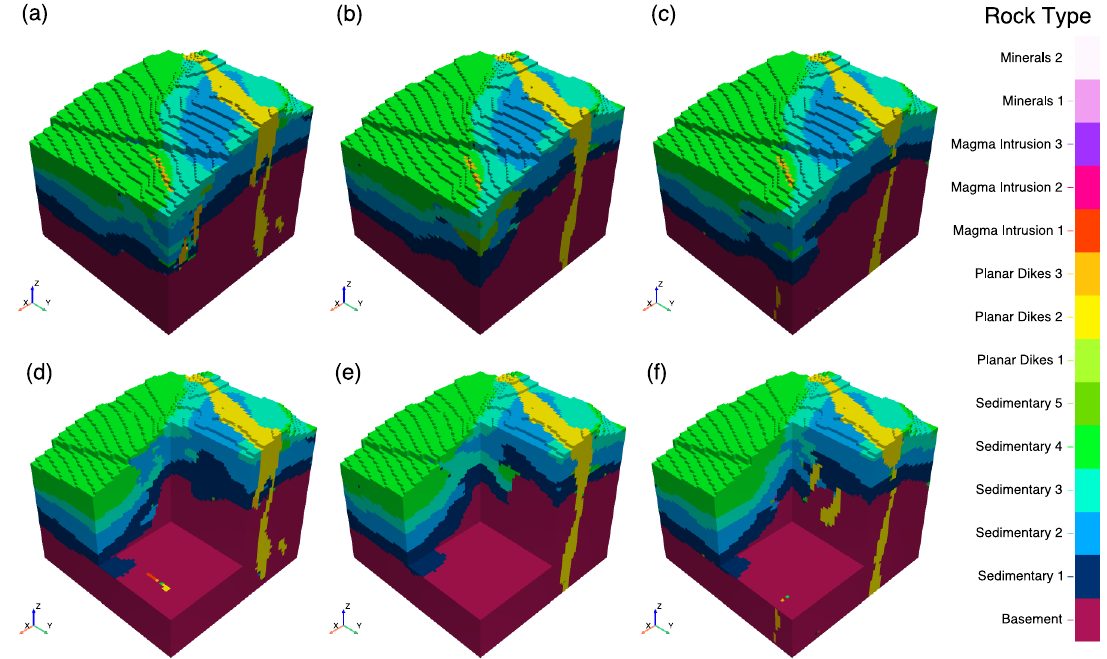} 
    \caption{Three geological models from $\pi_{\bfm | \bfd}$ via numerical integration conditioned on available lithological data (top row) and their corresponding cut-section views (bottom row). The models offer a complete 3D\add{ reconstruction with a resolution of} $(64 \times 64 \times 64)$\remove{ resolution reconstruction}, with the exterior surface of the models shown here for comparison. Cut-away views indicate changes throughout the internal range of the model, as reconstructions sample from a complex $\pi_{\bfm | \bfd}$ distribution.}
    \label{fig:recfrombh}
\end{figure}

Each instance of Gaussian noise\remove{d} maps to a different reconstruction conditioned on the sparse data, offering complete reconstructions of hypothetical yet probable geology, a technique referred to as inpainting. Three such volumetric realizations are shown in Figure~\ref{fig:recfrombh}.
While the models are identical at the ground surface where conditional information was fully specified, they diverge in areas\remove{ that were} not covered by the conditional data, for example, the extent and shape of the two planar dike features. 

\subsection{Conditional Feature Estimation and Reconstruction}

\begin{figure}[h]
  \centering
\includegraphics[width=\linewidth]{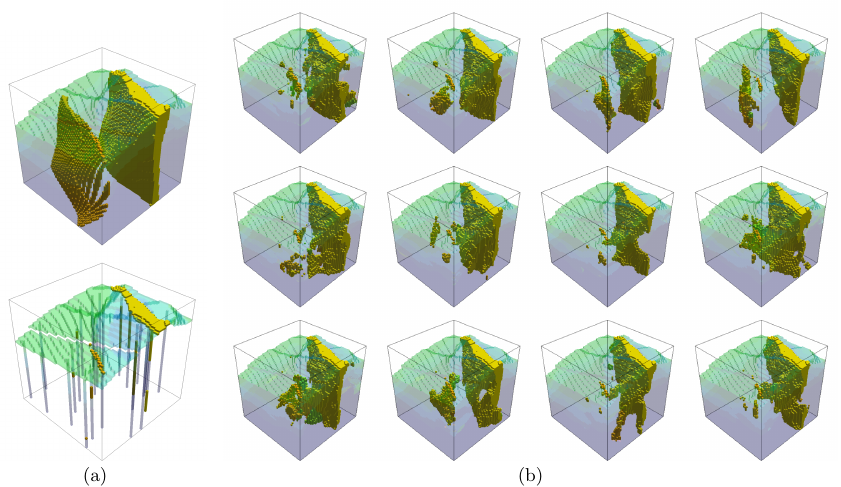}

  \caption{\textbf{Conditioned dike realizations.}
  All models are rendered with a transparent exterior to emphasize only the 3D lithology of the planar dikes.
  (a) True dike geometry in situ (yellow and orange) with sparse surface/borehole data.
  (b) Twelve conditional realizations sampled from the model. The reconstructions demonstrate the diverse sampling possible from a learned $\pi_{\bfm|\bfd}$ distribution. }
    \label{fig:dike_analysis}
\end{figure}

Individual features of a reconstructed model, such as dike formations, can be analyzed in greater detail using an ensemble of samples from the conditional distribution \change{$\pi_\bfm$}{$\pi_{\bfm | \bfd}$}. The planar dikes in the original model, the sparse dike data, and several reconstructions of their form are shown in Figure~\ref{fig:dike_analysis}.  The realizations show a high degree of variability due to the very limited conditional data. 

\begin{figure}[h]
  \includegraphics[width=\linewidth]{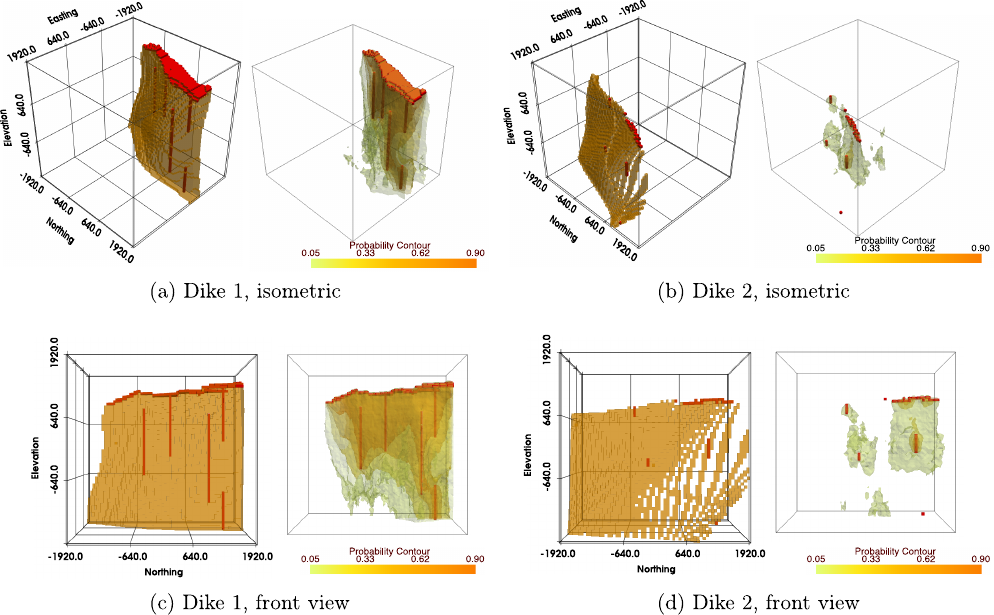}
    \caption{\textbf{Comparison of conditional dike realizations.} Two dike reconstructions from sparse data at different probability confidence intervals. In each subfigure, \remove{on the left is }the true dike\remove{, which} is shown \add{on the left }in gold, with sparse field data \change{shown}{indicated} in red. On the right is a probability estimate using an ensemble of 100 conditional reconstructions from the generative flow model. Four different probability thresholds are shown: $5\%$, $33\%$, $62\%$, and $90\%$.}
    \label{fig:dikeprob}
\end{figure}

A large ensemble of samples can be used as a probability estimate for the likelihood of any individual voxel being part of the larger dike structure. A pattern emerges when plotting the extent of the planar dike based on the contour shape of probabilistic confidence intervals. Shown in Figure~\ref{fig:dikeprob}, a lower confidence interval corresponds to a larger and more aggressive extrapolation range from the existing boreholes. The smaller dike shown in the model has insufficient conditional data to produce an accurate reconstruction, a problem exacerbated by the true extent of the dike being largely obscured by sediment\change{,}{;} see Figure~\ref{fig:model3d+data} (a) and Figure \ref{fig:dikeprob} (b) for a comparison of the small surface signature to the larger subsurface extent.

\section{Discussion}

Our results demonstrate the feasibility and potential of using generative artificial intelligence (AI) models, particularly flow matching techniques, to construct realistic and geologically plausible three-dimensional structural models. The work presented here integrates two key components. First, we construct a process for creating a large, synthetic, and geologically consistent dataset that seeks to embed the rich and complex knowledge of structural geology into a mathematical framework suitable for training machine learning models. Second, we train a generative AI model that implicitly learns both unconditional and conditional probability distributions of geological structures. By doing so, it efficiently captures essential characteristics of large\add{-}scale lithological units in the Earth’s crust and provides a highly complex prior to tackle the challenges common in geoscience applications, including data sparsity, uncertainty quantification, and the integration of diverse data modalities. Although the underlying simulation engine could, in principle, operate at finer scales or incorporate continuous physical properties such as conductivity, permeability, \add{and }density, our focus here is on modeling the categorical organization of major lithologic bodies conditioned by surface and borehole constraints.

\textbf{Advancements in Generative Modeling for Structural Geology.} Generative AI for structural geology offers a significant advantage over traditional deterministic approaches \change{through their}{due to its} capacity to define multiple plausible geological realizations, which is particularly valuable for\add{ addressing} the ill-posedness of geological reconstruction from often sparse and incomplete data. The use of synthetic data to train a generative model is an opportunity to specify prior knowledge that is vastly more complex than what would be permitted through classical techniques such as regularization. 
The unconditional models generated have a high degree of geological diversity, capturing variations in topography, layer geometry, and structural features such as dykes and intrusions. This type of synthetic modeling has numerous applications, including data augmentation for training machine learning models, benchmarking geophysical inversion algorithms, and enhancing geostatistical modeling techniques. Our results further reinforce previous studies that highlight the importance of generative approaches in geosciences \cite{wellmann2015PynoddyExperimentalPlatform, bergen2019MachineLearningDatadriven, woollam2021seisbenchtoolboxmachine}.

\textbf{Conditional Modeling, Reconstruction, and Uncertainty Quantification.} The ability to generate samples conditioned on sparse borehole data presents a major breakthrough for geological interpolation tasks. The technique can succeed where tools such as kriging often struggle, for example, when borehole data is limited or categorical. The generative model approach produces an ensemble of realizations that not only fit the observed data but also adhere to a probability distribution of plausible geology encoded into the synthetic training set.
The generated models are not only consistent with the conditional data but also diverse, \change{assisting}{supporting} tasks such as risk assessment and decision-making in resource exploration, subsurface engineering, and geological hazard analysis. By providing a suite of possible geological scenarios, our approach allows decision-makers to incorporate uncertainty into their models rather than relying on a single deterministic solution.

\textbf{Implications and Future Directions.} The ability to generate diverse, geologically plausible models has far-reaching implications for \change{a variety of}{various} geoscientific applications. One promising direction is the integration of our generative framework with geophysical inversion workflows, where subsurface properties inferred from geophysical data can be used to\remove{ further} constrain model realizations\add{ further}. Additionally, combining our approach with Bayesian inference techniques could enable probabilistic geological modeling, providing a more rigorous quantification of uncertainty. Future work could also extend this framework to finer resolutions or physical property domains where multiscale variability becomes essential.

\textbf{Conclusions.} The incorporation of generative AI into structural geology represents a paradigm shift, enabling more flexible, data-driven, and probabilistic interpretations of the Earth's subsurface. As these methods continue to evolve, they have the potential to transform geological modeling, supporting improved resource exploration \change{through, e.g., providing}{for example, by providing} novel types of regularization for geophysical inverse problems, as well as \add{enhancing} geohazard assessment and geotechnical decision-making.

Finally, while this study represents a significant step toward automation in structural geology, human expertise remains invaluable. Rather than replacing traditional geological interpretation, our approach should be viewed as a powerful tool that enhances geologists' ability to explore multiple hypotheses, validate models, and make informed decisions in the face of uncertainty.

\clearpage

%
%
\section*{Acknowledgments} For computer time, this research partially used the resources of the Supercomputing Core Laboratory at King Abdullah University of Science and Technology (KAUST).

\section*{Open Research}
The StructuralGeo simulation code (version~1.0) used to generate the synthetic geological models in this study is preserved at Zenodo \cite{StructuralGeo_v1} and \add{is }openly developed at GitHub (\url{https://github.com/eldadHaber/StructuralGeo/releases/tag/v1.0}), under a permissive open-source license.

The generative\change{-}{ }model training and evaluation code is preserved at Zenodo \cite{FlowTrain_v1} and developed at GitHub \begin{nolinenumbers} \newline
    (\url{https://github.com/chipnbits/flowtrain_stochastic_interpolation/releases/tag/v1.0.2}).
\end{nolinenumbers} This code supports reproducing both conditional and unconditional experiments.

The attention U-Net code and architecture are adapted from software by \citeA{wang_ddpm_pytorch} found at \url{https://github.com/lucidrains/denoising-diffusion-pytorch}.

All software is released under open-source licenses. No registration is required to access the data or software.


\section*{Author Contributions Statement}

S.G. developed the synthetic data\remove{}set generator and was involved in the generative AI model development. 
V.O. and S.Z. trained the foundation model.  
G.T. led the training and together with D.K. organized the KAUST-based effort. 
E.H. conceived of and led the overall project.
All authors participated in preparing and reviewing 
the manuscript. 

\textbf{Competing Interests Statement}: 
The authors declare no competing interests.

\textbf{Authorship and AI Tools}:
ChatGPT (GPT-5, OpenAI) was used in this manuscript to assist with spelling, grammar, and improvements to sentence and paragraph structure. The authors reviewed, verified, and approved all suggestions and take full responsibility for the content.


\bibliography{biblio}

%
%
%
%
%

\end{document}